\documentclass{article}

\usepackage{amsmath}
\usepackage{amssymb}
\usepackage{booktabs}
\usepackage{color}
\usepackage{graphicx}
\usepackage{listings}
\usepackage{microtype}
\usepackage{subcaption}
\usepackage{tikz}

\definecolor{maroon}{cmyk}{0, 0.87, 0.68, 0.32}
\definecolor{halfgray}{gray}{0.55}
\definecolor{ipython_frame}{RGB}{207, 207, 207}
\definecolor{ipython_bg}{RGB}{247, 247, 247}
\definecolor{ipython_red}{RGB}{186, 33, 33}
\definecolor{ipython_green}{RGB}{0, 128, 0}
\definecolor{ipython_cyan}{RGB}{64, 128, 128}
\definecolor{ipython_purple}{RGB}{170, 34, 255}

\usetikzlibrary{bayesnet}
\tikzstyle{obs_disc} = [rectangle, fill=gray!25, minimum size=15pt, draw=black]
\graphicspath{{figures/}}

\DeclareMathOperator*{\argmax}{arg\,max}
\DeclareMathOperator*{\argmin}{arg\,min}

\newcommand{\fbest}{f_{\text{best}}}

\newcommand{\dif}{\textup{d}}

\newcommand{\ff}{\mathbf{f}}

\newcommand{\kk}{{\mathbf{k}}}

\newcommand{\xx}{{\mathbf{x}}}

\newcommand{\zz}{{\mathbf{z}}}

\newcommand{\ftest}{f_\text{test}}
\newcommand{\xbest}{\xx_\text{best}}
\newcommand{\xnext}{\xx_\text{next}}

\newcommand{\cX}{\mathcal{X}}

\newcommand{\mK}{\mathsf{K}}

\newcommand{\RR}{\mathbb{R}}

\newcommand{\algref}[1]{{Algorithm~\ref{#1}}}
\newcommand{\figref}[1]{{Figure~\ref{#1}}}

\usepackage[nohyperref, accepted]{hill2019}
\icmltitlerunning{
Sampling Humans for Optimizing Preferences in Coloring Artwork
}

\begin{document}

\twocolumn[
\icmltitle{
Sampling Humans for Optimizing Preferences in Coloring Artwork
}

\icmlsetsymbol{equal}{*}
\begin{icmlauthorlist}
    \icmlauthor{Michael McCourt}{sigopt}
    \icmlauthor{Ian Dewancker}{kindred}
\end{icmlauthorlist}
\icmlaffiliation{kindred}{Kindred AI, Toronto, ON, CA}
\icmlaffiliation{sigopt}{SigOpt, San Francisco, CA, USA}
\icmlcorrespondingauthor{Michael McCourt}{mccourt@sigopt.com}

\icmlkeywords{Bayesian Optimization, Preference Learning}

\vskip 0.3in
]

\printAffiliationsAndNotice{}

\begin{abstract}
Many circumstances of practical importance have performance
or success metrics which exist implicitly---in the eye of the
beholder, so to speak.  Tuning aspects of such problems requires
working without defined metrics and only considering pairwise
comparisons or rankings.  In this paper, we review an existing
Bayesian optimization strategy for determining most-preferred
outcomes, and identify an adaptation to allow it to handle ties.
We then discuss some of the issues we have encountered when
humans use this optimization strategy to optimize coloring
a piece of abstract artwork.  We hope that, by participating
in this workshop, we can learn how other researchers encounter
difficulties unique to working with humans in the loop.
\end{abstract}

\section{Introduction}

Bayesian optimization is a sample-efficient strategy for
black-box optimization \cite{shahriari2016taking, frazier2018bayesian}.
In many practical circumstances, such as robotics systems, measuring performance is often complicated by having only an implicit understanding of utility, the existence of multiple competing metrics, or reliance on perceptual metrics that are not easily instrumented or quantified \cite{wirth2017survey, thatte2017sample, pinsler2018sample}.
In such circumstances, standard Bayesian optimization maybe be infeasible: it may be more practical to explicitly incorporate comparative feedback into the search for a stakeholder's \emph{most preferred}
outcome.  Preference-based optimization offers an approach that relies only on pairwise comparative evaluations, rather than forcing the design of a single criteria or utility for which standard black-box optimization can be applied \cite{cano2018automatic, burger2017automated}.

This strategy of preference-based optimization, often involving a human-in-the-loop, has been developed over the past decade
\cite{brochu2010bayesian,eric2008active, gonzalez2017preferential,
houlsby2012collaborative,thatte2017sample}.
These tools build on the foundation of sequential model-based Bayesian optimization
\cite{bergstra2011algorithms,hutter2011sequential,
snoek2012practical}.
The structure of these optimization strategies is to sequentially query the user with (generally) two possible options, of which
the user must choose one.

In this article, we start from an existing preference model
defined by a Gaussian process latent variable model; we
describe an extension, first presented in \cite{dewancker2016}
to allow for the user to report
configurations as equivalently preferable.
\begin{figure}[ht]
  \begin{center}
  \includegraphics[width=0.4\textwidth]{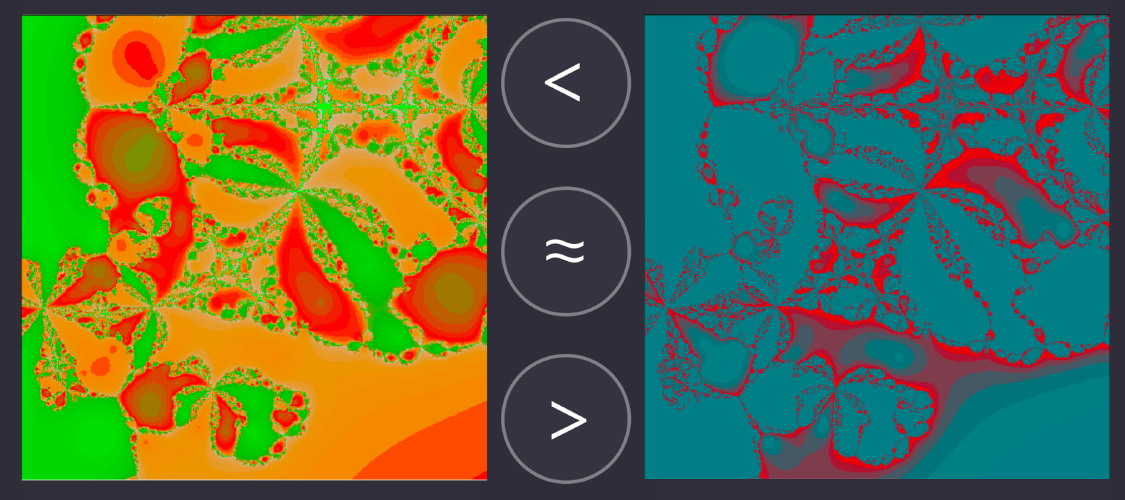}
  \caption{
    Preference-based optimization of artwork coloring. A user is asked to compare a sequence of configurations of a planning system as either worse, better or equivalent.  The pairwise comparisons are used to refine the search for the optimal configuration.
    \label{fig:planning2pics}
  }    
  \end{center}
\end{figure}
These so-called ``ties'' provide the opportunity for users
to avoid stating a preference in circumstances
where such imprecision exists; we present a short empirical
analysis to consider the impact of allowing ties.
We also show a synthetic example for this strategy to
identify a user's most preferred Pareto efficient outcomes
without exploring the entire efficient frontier
\cite{ehrgott2005multicriteria, knowles2006parego}.

Our eventual goal for this work is to be able to identify
most-preferred outcomes for circumstances which lack quantifiable
metrics.  We use coloring artwork to define one such
circumstances as a stand-in for application-specific or
confidential circumstances.  We present the desired
experimental workflow, and also some of the complications
which have occurred in initial testing.  We hope that
participating in this ICML Human in the Loop workshop
will introduce us to others who have insights in the
psychological complications involved in incorporating humans
into similar training/tuning/iterative processes.

\vspace{-2mm}
\section{A Preference Model Supporting Ties
\label{sec:gen_inst}}

The use of Gaussian process latent variable models (GPLVM)
for capturing user preferences has been well studied in the past
\cite{brochu2010bayesian, eric2008active,
chu2005preference, guo2010Gaussian}.
Previous preference models have required that the user
consider two alternatives $\xx^{(1)},\xx^{(2)}\in\Omega\subset\RR^D$
and state a binary preference.
However, even experts occasionally have
difficulty discerning two alternatives in terms of absolute
preference.  To this end, we extend the discrete
preference observations with an additional third option: 
specifying equivalence between the two alternatives.
Specifically, we adopt a modified Bradley-Terry model that
supports ties, or configurations with equivalent preference
\cite{rao1967ties}; summarized in
\figref{fig:graphicalmodel}.

\begin{figure}[ht]
	\centering
	\begin{tikzpicture}
	
	\node[obs_disc]		          (c)     {$c$};
	\node[latent, left=1.65cm of c]         (f)     {$\ff$};
	\node[obs, above=1.5cm of f]            (X)     {$\cX$};
	\factor[above=0.65cm of f] {K} {right:$\mK$} {}{}; %
	
	\factor[left=0.7cm of c]          {z}     {below:$\pmb{\pi}$} {}{};
	
	\factor[left=0.55cm of K] {theta} {below:$\pmb{\theta}$} {}{};
	\node[latent, left=1.25cm of K]         (gamma) {$\pmb{\gamma}$};
	\node[const, above=0.78cm of z]        (beta)  {$\beta$};
	\node[const, above=0.90cm of theta]      (alpha) {$\pmb{\alpha}$};
	\edge [] {z} {c} ; %
	\edge {f} {z} ; %
	\edge {theta} {K} ; %
	\edge {gamma} {theta} ; %
	\edge {X} {K} ; %
	\edge {K} {f} ; %
	\edge {beta}  {z} ; %
	\edge {alpha} {theta} ; %
	
	\plate {zc} {(z)(c)} {$M$} ;
	\end{tikzpicture}
	
	\vspace{-5mm}
    \begin{align*}
    	\gamma_d &\sim \textbf{Normal}(0, 1)   \\
    	\theta_d &= S(\gamma_d)(\alpha_{U, d} - \alpha_{L, d}) + \alpha_{L, d} \\
    	(\mathbf{K})_{i,j} &= rbf(\mathbf{x}_{i}, \mathbf{x}	_{j}, \pmb{\theta}	) \\
    	\mathbf{f} &\sim \textbf{MVNormal}(\mathbf{0}, \mathbf{K}) \\
    	\xi^{(1)}_m &= S((f^{(1)}_{m} - f^{(2)}_{m}) / \sqrt{2\sigma^2}), \ \ \xi^{(2)}_m = 1 - \xi^{(1)}_m\ \\
    	\pi^{\prec}_m &= \frac{\xi^{(2)}_m}{\xi^{(2)}_m\!+\!\beta \xi^{(1)}_m}, \ \pi^{\approx}_m = \frac{(\beta^2-1)\xi^{(1)}_m\xi^{(2)}_m}{(\xi^{(1)}_m\!+\!\beta \xi^{(2)}_m)(\xi^{(2)}_m\!+\!\beta \xi^{(1)}_m)} \\
    	c_m &\sim \textbf{Categorical}_3(\pi^{\prec}_m, \ \pi^{\approx}_m, \ (1 - \pi^{\prec}_m - \pi^{\approx}_m))
    \end{align*}
    
    \vspace{-3mm}
	\caption{
		\label{fig:graphicalmodel}
		Graphical model and associated distributions
		of GPLVM for preferences including ties.
	}
\end{figure}

After $M$ such preferences have been stated,
we denote the results as
$\{(\xx^{(1)}_m,\xx^{(2)}_m,c_m)\}_{m=1}^M$, where
$c_m\in\Xi$ for $\Xi=\{<,>,\approx\}$.  Among all the $M$
queries, there are $N$ unique $\xx$ locations:
$\cX = \{\xx_1,\ldots,\xx_N\}$.  

The model draws latent function vectors $\mathbf{f} \in \RR^N$ from a Gaussian
process prior where each entry
corresponds to one of the $N$ unique query points $(\cX)$ the user has compared.  The $\pmb{\gamma}\in\RR^D$ variables are drawn from normal priors and then transformed to form the lengthscales of the 
covariance function, used to populate $\mathbf{K}$, the covariance matrix.

Each $\gamma_d$ variable is transformed to always lie within the bounds of the lengthscales, 
specified by $\pmb{\alpha}_U$ and $\pmb{\alpha}_L$, and produce a vector of lengthscales $\pmb{\theta}$.
Here $S(x) = \frac{1}{1 + e^{-x}}$ and $rbf(\mathbf{x}_i,\mathbf{x}_j,\pmb{\theta}) = \sigma^2 \exp\Big(-\frac{1}{2} \sum_{d=1}^D \frac{1}{\theta_d^2} (x_{i,d} - x_{j,d})^2 \Big)$.

The generalized Bradley-Terry model \cite{rao1967ties} relates the observed discrete preference data $c_m$ to the latent function values $(f^{(1)}_m, f^{(2)}_m)$ associated with the two points $(\mathbf{x}^{(1)}_m, \mathbf{x}^{(2)}_m)$ compared 
by the user during an interactive query.  The tie parameter $\beta \geq 1$, is inversely related to the precision
with which a user can state a preference.  A higher value for $\beta$ leads to more mass being placed in the equivalence bin $(\pi^{\approx}_m)$ of the categorical distribution over the three
possible preference outcomes for two query points.

\subsection{Variational Inference}

In place of approximating the posterior with a multivariate Gaussian using the Laplace approximation around a MAP estimate of the latent variables \cite{chu2005preference,guo2010Gaussian}, we 
opt for an approximation that employs variational inference.  We set out to 
approximate $p( \mathbf{z} \ | \ \cX, \mathbf{c})$, the posterior of the latent random variables, where $\mathbf{z} = \{\mathbf{f}, \pmb{\gamma}\}$ is the combined set of latent random variables in our model.

We use a mean field approximation strategy to construct our approximating distribution $q$ : a factored set of Gaussians each parameterized by a mean and variance as shown below.
\vspace{-2mm}
\begin{align*}
 p( \mathbf{z} |  \cX, \mathbf{c}) \approx q(\mathbf{z}  ;  \pmb{\lambda} ) &= \prod_{i=1}^{N} \mathcal{N}(z_i | {\lambda_{\mu}^i},  {\lambda_{\sigma}^i}) \prod_{d=1}^{D} \mathcal{N}(z_d  |  {\lambda_{\mu}^d},  {\lambda_{\sigma}^d})
\end{align*}
We rely on black box variational inference techniques \cite{tran2016edward,ranganath2013black} to perform the optimization required to recover the variational parameters $\pmb{\lambda}$ that minimize the reverse KL divergence between
the true posterior distribution $p$ and the approximating distribution $q$.
In total there will be $2N + 2D$ variational parameters; two for each of the $N$ entries in $\mathbf{f}$ and two for each of the $D$ elements of $\pmb{\gamma}$.
The posterior inference problem is transformed into a minimization of a tractable expected value \cite{tran2016edward,ranganath2013black}.
\vspace{1mm}
\begin{align*}
\pmb{\lambda}^* &= \argmin_{\pmb{\lambda}} \ KL(q \ || \ p) \\
 &= \argmin_{\pmb{\lambda}} \int q(\mathbf{z} \ ; \ \pmb{\lambda}) \ \text{log} \ \frac{q(\mathbf{z} \ ; \ \pmb{\lambda})}{p( \mathbf{z}, \cX, \mathbf{c})} \, \dif\mathbf{z} \\
 &= \argmin_{\pmb{\lambda}} \ \mathbb{E}_{\mathbf{z} \sim q(\mathbf{z} ; \pmb{\lambda})} \left[ \text{log} \ q(\mathbf{z} \ ; \ \pmb{\lambda}) - \text{log} \ p( \mathbf{z}, \cX, \mathbf{c} ) \right] 
\end{align*}

\subsection{Sequential Preference Based Optimization\label{sec:pbo}}

To determine the next point, $\xnext$ to be presented to the user as comparison point, we adopt a strategy that searches the domain for where the expected improvement of latent function  
    is highest relative to the current, most preferred point ($\xbest$) \cite{eric2008active}.  With our approximation $q(\zz \ ; \ \pmb{\lambda} )$  
of the posterior of the latent variables of the preference model, it is possible to explore the use of an integrated acquisition function, as proposed in \cite{snoek2012practical}. 
We can produce a Monte Carlo estimate of the integrated expected improvement.
\begin{align*}
    \kk(\xx)^\top &= [rbf(\xx,\xx_1, \pmb{\theta}) \ \cdots \ rbf(\xx,\xx_N, \pmb{\theta}) ] \\
    \mu(\xx) &= \kk(\xx)^\top\mK^{-1} \ff \\
    s(\xx)^2 &= rbf(\xx,\xx, \pmb{\theta}) - \kk(\xx)^\top \mK^{-1}\kk(\xx) \\
    \nu &= (\mu(\xx) - f_{best}) / s(\mathbf{x}) \\
    a_{\mathsf{EI}}(\xx ; \zz  ) &= \begin{cases}
    s(\xx)(\nu\Phi(\nu) + \phi(\nu)),& \hspace{2mm} \text{if} \ s(\mathbf{x}^*) > 0\\
    0 ,& \hspace{2mm} \text{if} \ s(\mathbf{x}^*) = 0
    \end{cases} \\
    \xnext &= \argmax_{\xx\in\Omega} \int a_{\mathsf{EI}}(\xx ; \zz  ) q(\zz \ ; \ \pmb{\lambda} ) \,\dif\mathbf{z}
\end{align*}

Here $\Phi(\cdot)$ and $\phi(\cdot)$ denote the CDF and PDF of
the standard normal distribution respectively.  The value $\fbest$ is the latent function value associated with the currently most preferred configuration $\xbest$.
The user is always asked to compare $\xnext$ against the current most preferred point $\xbest$, and the most preferred point is updated as a result of this comparison.  \algref{prefopt_algo} encapsulates the sequential optimization process using discrete preference observations.

\begin{algorithm}
\caption{Sequential Preference-Based Optimization}
\label{prefopt_algo}
\begin{algorithmic}
		\STATE {\bfseries Input:}  $\Omega$ 
		\STATE $\mathcal{D},\mathbf{x}_{\text{best}}  \gets \textsc{InitUserPrefs}( \Omega )$
		\FOR{$i \gets 1$ {\bfseries to} $T$}
		\STATE $ q(\mathbf{z} \ ; \ \pmb{\lambda} ) \approx p( \mathbf{z} \ | \ \mathcal{D}) $
		\STATE $ \xnext  \gets \argmax_{\xx \in \Omega} \, \int a_{\mathsf{EI}}(\xx ; \zz  ) q(\zz \ ; \ \pmb{\lambda} ) \,\dif\zz$  
		\STATE $ c_i \gets \textsc{GetUserPref}(\xbest, \xnext)$
		\STATE $ \mathcal{D} \gets \mathcal{D} \, \cup \ (\xbest, \xnext, c_i)$
		\IF{$c_i$ == $<$}
			\STATE $ \xbest \gets \xnext $
		\ENDIF	
		\ENDFOR
\end{algorithmic}
\end{algorithm}

Here, we initialize the search with $2D+1$ samples from a latin hypercube sequence.
We fix the quantities $\beta=1.1$, $\sigma=0.1$,
and we set
$\alpha_{L, d}=.01r$ and $\alpha_{U, d}=5r$ where $r$ is
the length of the optimization domain in the $d$th dimension.

\subsection{Synthetic Numerical Experiments Involving Ties}
\begin{figure}[ht]
    \centering
    \begin{subfigure}{\linewidth}
        \centering
        \includegraphics[width=.7\linewidth]{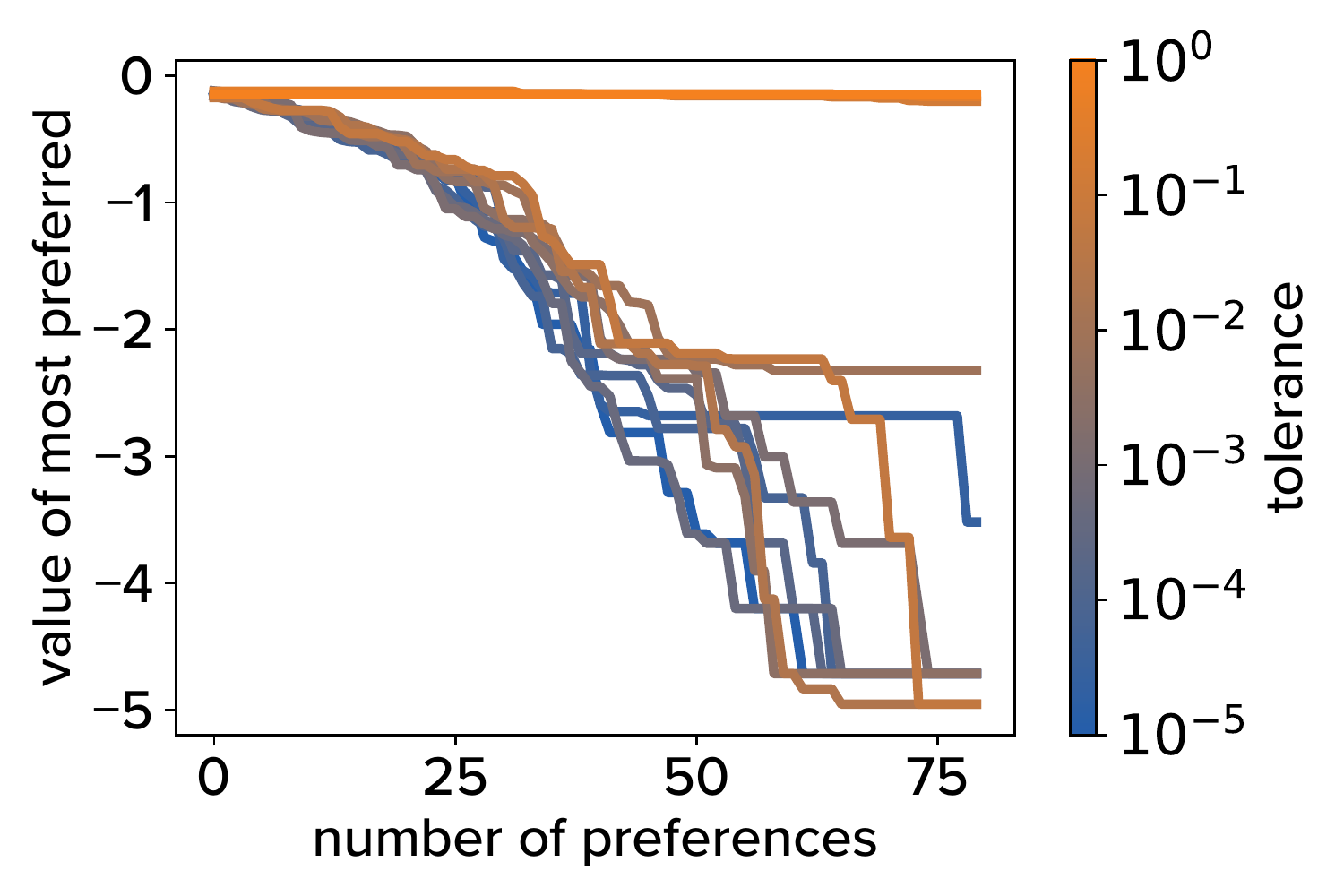}
    	\caption{
    		Medians of the trials are plotted as a function
    		of the number of preferences sampled.
    	}
    	\label{fig:shekel_color_converge}
	\end{subfigure}
	\hfill
    \begin{subfigure}{\linewidth}
        \centering
        \includegraphics[width=.7\linewidth]{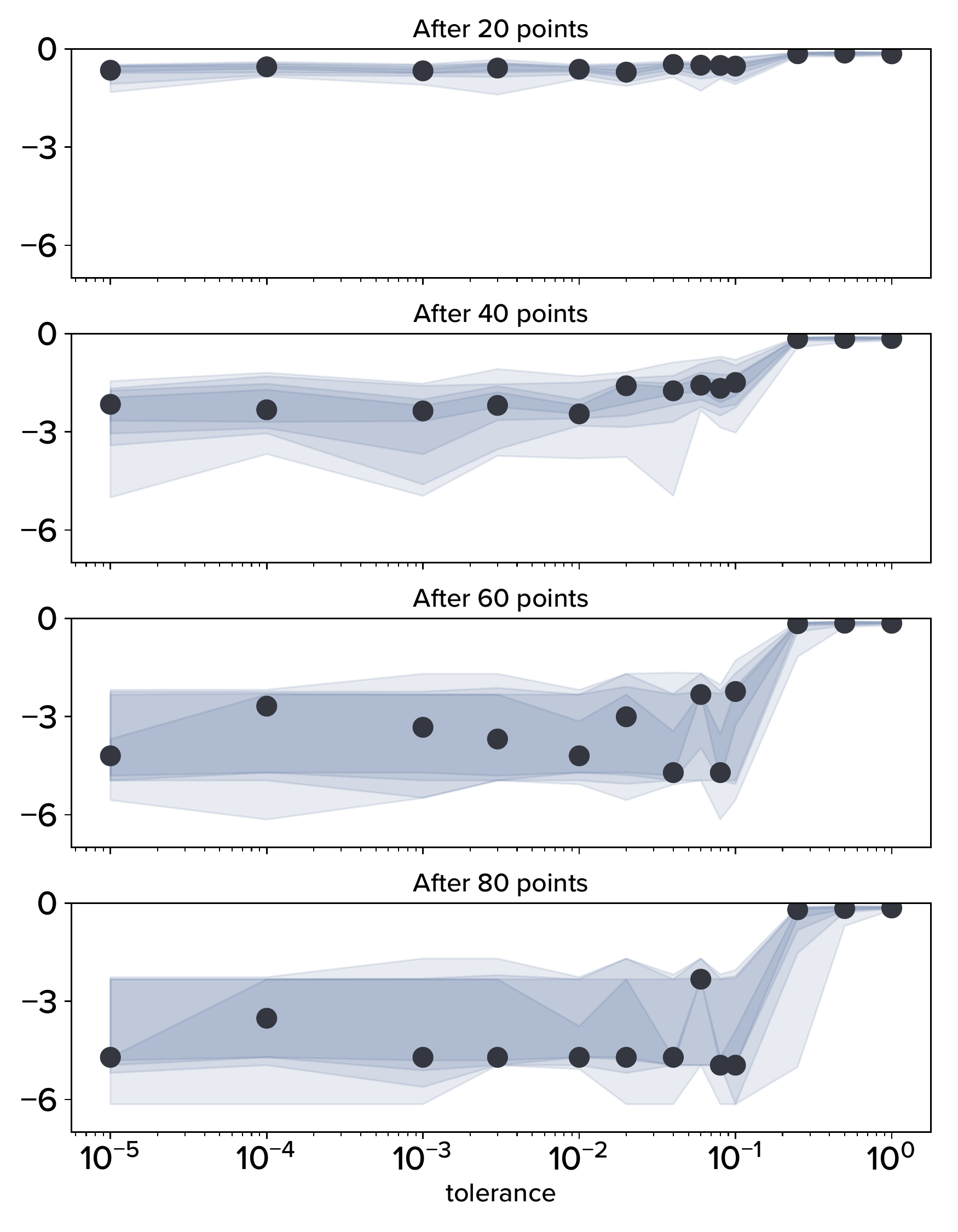}
    	\caption{
    		Interpercentile ranges (from darkest [40-60]
    		to lightest [10-90]) of the trials are presented
    		as a function of tolerance.  The medians are plotted
    		with charcoal circles.
    	}
        \label{fig:shekel_color_progress}
	\end{subfigure}
	\caption{
		\label{fig:shekel_color}
		The convergence behavior of \algref{prefopt_algo}
		as tested over 20 independently trials on the
		\texttt{Shekel05} function.
		While large tolerances seem to stagnate,
		sufficiently small values (below $~0.1$)
		move towards the minimum.  Also, the lowest
		tolerance does not consistently perform better.
	}
\end{figure}

\begin{figure*}[ht] 
    \begin{center}
    \includegraphics[width=.4\linewidth]{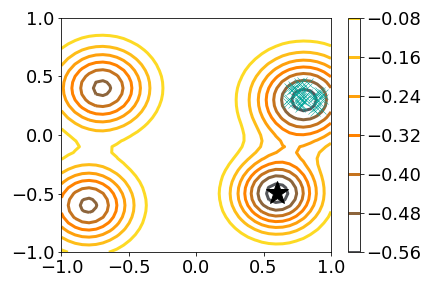}
    \includegraphics[width=.4\linewidth]{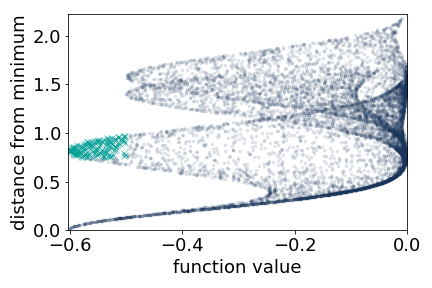}
    \includegraphics[width=.3\linewidth]{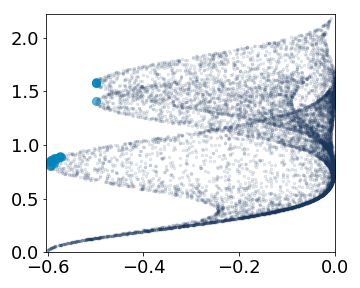}
    \includegraphics[width=.3\linewidth]{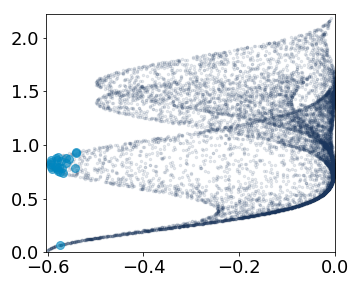}
    \includegraphics[width=.3\linewidth]{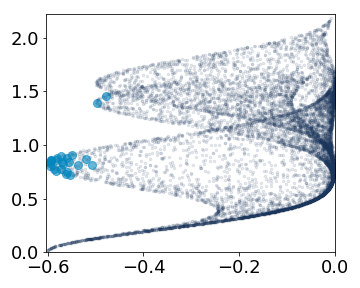}
	\caption{
		\label{fig:mc_utility}
		Results from different tolerances considered in
		a multi-objective setting -- trying to minimize a function
		while being far from the true minimum.
		\emph{(top left)}: The function, with a black $\star$
		for the minimum and green $\times$ for the preferred region.
		\emph{(top right)}: The feasible region showing the
		possible function values and their associated distance
		from the minimum; the green $\times$ represent the preferred region.
		\emph{(bottom)}: The final outcome found from each of 20
		independent trials for tolerance 0.003 (left), 0.03
		(middle) and 0.1 (right).
	}
  \end{center}
\end{figure*}

We now experiment to understand the impact of allowing
ties into the optimization process.  To do so, we phrase
the preference optimization process using a scalar function
$\ftest$ and a stated tolerance $\epsilon$,
\begin{align}
    \begin{cases}
        \xx^{(1)} \approx \xx^{(2)} , \hspace{2mm} \text{if}, \ \ |\ftest(\xx^{(1)}) - \ftest(\xx^{(2)})| \leq \epsilon \\
        \xx^{(1)} \succ \xx^{(2)} , \hspace{2mm} \text{else if}, \ \ftest(\xx^{(1)}) < \ftest(\xx^{(2)}) \\
        \xx^{(1)} \prec \xx^{(2)} , \hspace{2mm} \text{otherwise},
    \end{cases} 
    \label{eq:prefdef}
\end{align}

In a first experiment, we consider a standard scalar test function
as our $\ftest$, the \texttt{Shekel05} function
\cite{mccourt_test}.  We analyze the convergence behavior of
\algref{prefopt_algo} on this function with a range of tolerance
values between $10^{-5}$ and 1.
\figref{fig:shekel_color_converge} shows the convergence behavior
as dependent on tolerance.

\figref{fig:mc_utility} shows the impact of varying tolerances on
a multicriteria optimization problem using a multi-modal
function in two dimensions.  Success is defined as trying to
reach the region containing the local minimum with the second
lowest value; we approach this by balancing the competing
objectives of minimizing function value and maximizing distance
from the global minimum.
To automate the testing process (pretend that a human was in the
loop) we define a nonlinear scalarized quality function
\[
    \ftest(\xx) = [
        4(f(\xx) + .6)^3 +
        .3(\|\xx-\xx_*\| - 2.4)^3
    ]^{1/3},
\]
where $\xx_*$ is the true minimum.
As we can see in the bottom
portion of \figref{fig:mc_utility}, the clearest impact of
tighter tolerances seems to be the narrowing down of results
to be closer to the Pareto frontier.

\section{Proposed Artwork Coloring Examples\label{sec:artworkcoloring}}

The eventual goal of our preference-based optimization
strategy is to be able to identify most-preferred outcomes
when a scalar function such as $\ftest$ does not exist.
We now present a testing situation which lacks any such
$\ftest$: the coloring of an abstract piece of art.  The
shape in particular is shown in \figref{fig:planning2pics}
with two possible colorings presented, from which the
user is expected to choose between them, or that they 
are both ``roughly equal''.

The shape is a fractal, defined in \cite{EricCode}, to
have 3 polynomial roots which points in the complex
domain converge to through Newton's method at
different rates.
The search domain for this
coloring problem is, at present, a 10 dimensional space
defining 3 different colors associated with the 3 polynomial
roots, transition speed between those 3 colors, and the base
color (which has an effect of dulling the colors).

Our goal in running these experiments was to determine the
performance of \algref{prefopt_algo} relative to a purely
random search \cite{bergstra2012random} of the domain.
In particular, users would be asked to run through some
number of optimizations using random search and some number
using \algref{prefopt_algo} and then a final comparison
asking which would be preferred between the most-preferred
from the two strategies.

\subsection{Initial Testing and User Feedback}

We built a webapp to provide a sequence of comparisons to
users in the format of \figref{fig:planning2pics}; the
incumbent most-preferred is always on the right-hand side.
This was the first point of feedback that we received from
users: ``Why does the winner always stick around?''  While
it presents a viable optimization strategy, we may need
to be more cognizant of the persistent presence of the
$\xbest$ during the optimization.

Additionally, some users complained that they felt ``tricked''
or that we ``misled'' them when learning that some of
the optimizations were powered by random search rather than
something more intelligent/adaptive.  This was surely caused
by running tests on people who had already been informed
of the purpose of the testing (a form of the placebo effect,
perhaps).  Even though testers did not know whether they
were running with random search or preference optimization
(to avoid a true placebo effect), some frustration around
the distinction was present.

A more severe complication, and one which we had failed to
consider, is the user's sense of \emph{completion}.  In
the pure \algref{prefopt_algo}, a budget $T$ would be set
at the start.  It seemed, though, that users were frustrated
by this fixed budget.  Some commented "Do I keep going?" or
"I think I'm done." part way through the budget; others said
"I want to keep going." after reaching the budget.

\subsection{Proposed Next Steps}

One option to address the user discomfort with always seeing
$\xbest$ is comparisons is to consider more of a
tournament strategy.  While this is straightforward when randomly
searching the space, it is more complicated to adapt
\algref{prefopt_algo} for this situation.

To address the user's discomfort in the process of random
versus intelligent testing, we should consider not informing
people how the testing process is designed and proceeding.
Originally, we were afraid that testers would not find this
opaque presentation reasonable, but it may be preferable to
not give users any understanding as to the structure.  One
user commented
``I wasn't necessarily choosing what I thought was the most
aesthetically pleasing but making choices to try and direct
the system in a certain way that would generate better images
down the road.'' which is, obviously, not part of our
optimization strategy.

The frustration users felt around the budget is reasonable,
but not something that we immediately know how to address.
One option might be to allow users to run the optimization
until they ``feel'' it is complete; doing so changes the
experimental structure but might yield additional insights
into user behavior (especially regarding their tolerance).

Probably the most problematic component of this testing
framework was something alluded to by users but not
explicitly stated: a lack of understanding about the actual
range of outcomes in the search space.  Users asked
"Is this how it is supposed to look?" and "What colors are
available?" during the testing, which led us to realize
that, without any sense of what could occur in the coloring,
it would be difficult to confidently make judgments.

It is our hope that we can discuss these final two points
with other researchers at this ICML HILL workshop and try to
come away with strategies for future experimentation.

\section*{Acknowledgements}

Special thanks to Eric Lee for helping prepare the graphics
for the preference comparison experiments.


\begin{thebibliography}{25}
	\providecommand{\natexlab}[1]{#1}
	\providecommand{\url}[1]{\texttt{#1}}
	\expandafter\ifx\csname urlstyle\endcsname\relax
	\providecommand{\doi}[1]{doi: #1}\else
	\providecommand{\doi}{doi: \begingroup \urlstyle{rm}\Url}\fi
	
	\bibitem[Bergstra \& Bengio(2012)Bergstra and Bengio]{bergstra2012random}
	Bergstra, J. and Bengio, Y.
	\newblock Random search for hyper-parameter optimization.
	\newblock \emph{Journal of Machine Learning Research}, 13\penalty0
	(Feb):\penalty0 281--305, 2012.
	
	\bibitem[Bergstra et~al.(2011)Bergstra, Bardenet, Bengio, and
	K{\'e}gl]{bergstra2011algorithms}
	Bergstra, J.~S., Bardenet, R., Bengio, Y., and K{\'e}gl, B.
	\newblock Algorithms for hyper-parameter optimization.
	\newblock In \emph{Advances in Neural Information Processing Systems}, pp.\
	2546--2554, 2011.
	
	\bibitem[Brochu et~al.(2008)Brochu, de~Freitas, and Ghosh]{eric2008active}
	Brochu, E., de~Freitas, N., and Ghosh, A.
	\newblock Active preference learning with discrete choice data.
	\newblock In \emph{Advances in Neural Information Processing Systems}, pp.\
	409--416, 2008.
	
	\bibitem[Brochu et~al.(2010)Brochu, Brochu, and de~Freitas]{brochu2010bayesian}
	Brochu, E., Brochu, T., and de~Freitas, N.
	\newblock A bayesian interactive optimization approach to procedural animation
	design.
	\newblock In \emph{Proceedings of the 2010 ACM SIGGRAPH/Eurographics Symposium
		on Computer Animation}, pp.\  103--112. Eurographics Association, 2010.
	
	\bibitem[Burger et~al.(2017)Burger, Bharatheesha, van Eert, and
	Babu{\v{s}}ka]{burger2017automated}
	Burger, R., Bharatheesha, M., van Eert, M., and Babu{\v{s}}ka, R.
	\newblock Automated tuning and configuration of path planning algorithms.
	\newblock In \emph{2017 IEEE International Conference on Robotics and
		Automation (ICRA)}, pp.\  4371--4376. IEEE, 2017.
	
	\bibitem[Cano et~al.(2018)Cano, Yang, Bodin, Nagarajan, and
	O'Boyle]{cano2018automatic}
	Cano, J., Yang, Y., Bodin, B., Nagarajan, V., and O'Boyle, M.
	\newblock Automatic parameter tuning of motion planning algorithms.
	\newblock In \emph{2018 IEEE/RSJ International Conference on Intelligent Robots
		and Systems (IROS)}, pp.\  8103--8109. IEEE, 2018.
	
	\bibitem[Chu \& Ghahramani(2005)Chu and Ghahramani]{chu2005preference}
	Chu, W. and Ghahramani, Z.
	\newblock Preference learning with gaussian processes.
	\newblock In \emph{Proceedings of the 22nd International Conference on Machine
		Learning}, pp.\  137--144. ACM, 2005.
	
	\bibitem[Dewancker et~al.(2016)Dewancker, McCourt, and
	Ainsworth]{dewancker2016}
	Dewancker, I., McCourt, M., and Ainsworth, S.
	\newblock Interactive preference learning of utility functions for
	multi-objective optimization.
	\newblock In \emph{NIPS Future of Interactive Learning Machines Workshop},
	2016.
	
	\bibitem[Ehrgott(2005)]{ehrgott2005multicriteria}
	Ehrgott, M.
	\newblock \emph{Multicriteria optimization}, volume 491.
	\newblock Springer Science \& Business Media, 2005.
	
	\bibitem[Frazier(2018)]{frazier2018bayesian}
	Frazier, P.~I.
	\newblock Bayesian optimization.
	\newblock In \emph{Recent Advances in Optimization and Modeling of Contemporary
		Problems}, pp.\  255--278. INFORMS, 2018.
	
	\bibitem[Gonzalez et~al.(2017)Gonzalez, Dai, Damianou, and
	Lawrence]{gonzalez2017preferential}
	Gonzalez, J., Dai, Z., Damianou, A., and Lawrence, N.~D.
	\newblock Preferential bayesian optimization.
	\newblock \emph{arXiv preprint arXiv:1704.03651}, 2017.
	
	\bibitem[Guo et~al.(2010)Guo, Sanner, and Bonilla]{guo2010Gaussian}
	Guo, S., Sanner, S., and Bonilla, E.~V.
	\newblock Gaussian process preference elicitation.
	\newblock In \emph{Advances in Neural Information Processing Systems}, pp.\
	262--270, 2010.
	
	\bibitem[Houlsby et~al.(2012)Houlsby, Huszar, Ghahramani, and
	Hern{\'a}ndez-Lobato]{houlsby2012collaborative}
	Houlsby, N., Huszar, F., Ghahramani, Z., and Hern{\'a}ndez-Lobato, J.~M.
	\newblock Collaborative gaussian processes for preference learning.
	\newblock In \emph{Advances in Neural Information Processing Systems}, pp.\
	2096--2104, 2012.
	
	\bibitem[Hutter et~al.(2011)Hutter, Hoos, and
	Leyton-Brown]{hutter2011sequential}
	Hutter, F., Hoos, H.~H., and Leyton-Brown, K.
	\newblock Sequential model-based optimization for general algorithm
	configuration.
	\newblock In \emph{Learning and Intelligent Optimization}, pp.\  507--523.
	Springer, 2011.
	
	\bibitem[Knowles(2006)]{knowles2006parego}
	Knowles, J.
	\newblock {ParEGO}: a hybrid algorithm with on-line landscape approximation for
	expensive multiobjective optimization problems.
	\newblock \emph{IEEE Transactions on Evolutionary Computation}, 10\penalty0
	(1):\penalty0 50--66, 2006.
	
	\bibitem[Lee(2014)]{EricCode}
	Lee, E.~H.
	\newblock Alternating newton's method.
	\newblock \small \url{github.com/ericlee0803/NewtonEllipsoid}, 2014.
	
	\bibitem[McCourt(2016)]{mccourt_test}
	McCourt, M.
	\newblock {O}ptimization {T}est {F}unctions.
	\newblock \url{github.com/sigopt/evalset}, 2016.
	
	\bibitem[Pinsler et~al.(2018)Pinsler, Akrour, Osa, Peters, and
	Neumann]{pinsler2018sample}
	Pinsler, R., Akrour, R., Osa, T., Peters, J., and Neumann, G.
	\newblock Sample and feedback efficient hierarchical reinforcement learning
	from human preferences.
	\newblock In \emph{2018 IEEE International Conference on Robotics and
		Automation (ICRA)}, pp.\  596--601. IEEE, 2018.
	
	\bibitem[Ranganath et~al.(2013)Ranganath, Gerrish, and
	Blei]{ranganath2013black}
	Ranganath, R., Gerrish, S., and Blei, D.~M.
	\newblock Black box variational inference.
	\newblock \emph{arXiv preprint arXiv:1401.0118}, 2013.
	
	\bibitem[Rao \& Kupper(1967)Rao and Kupper]{rao1967ties}
	Rao, P. and Kupper, L.~L.
	\newblock Ties in paired-comparison experiments: A generalization of the
	bradley-terry model.
	\newblock \emph{Journal of the American Statistical Association}, 62\penalty0
	(317):\penalty0 194--204, 1967.
	
	\bibitem[Shahriari et~al.(2016)Shahriari, Swersky, Wang, Adams, and
	De~Freitas]{shahriari2016taking}
	Shahriari, B., Swersky, K., Wang, Z., Adams, R.~P., and De~Freitas, N.
	\newblock Taking the human out of the loop: A review of bayesian optimization.
	\newblock \emph{Proceedings of the IEEE}, 104\penalty0 (1):\penalty0 148--175,
	2016.
	
	\bibitem[Snoek et~al.(2012)Snoek, Larochelle, and Adams]{snoek2012practical}
	Snoek, J., Larochelle, H., and Adams, R.~P.
	\newblock Practical bayesian optimization of machine learning algorithms.
	\newblock In \emph{Advances in Neural Information Processing Systems}, pp.\
	2951--2959, 2012.
	
	\bibitem[Thatte et~al.(2017)Thatte, Duan, and Geyer]{thatte2017sample}
	Thatte, N., Duan, H., and Geyer, H.
	\newblock A sample-efficient black-box optimizer to train policies for
	human-in-the-loop systems with user preferences.
	\newblock \emph{IEEE Robotics and Automation Letters}, 2\penalty0 (2):\penalty0
	993--1000, 2017.
	
	\bibitem[Tran et~al.(2016)Tran, Kucukelbir, Dieng, Rudolph, Liang, and
	Blei]{tran2016edward}
	Tran, D., Kucukelbir, A., Dieng, A.~B., Rudolph, M., Liang, D., and Blei, D.~M.
	\newblock {Edward: A library for probabilistic modeling, inference, and
		criticism}.
	\newblock \emph{arXiv preprint arXiv:1610.09787}, 2016.
	
	\bibitem[Wirth et~al.(2017)Wirth, Akrour, Neumann, and
	F{\"u}rnkranz]{wirth2017survey}
	Wirth, C., Akrour, R., Neumann, G., and F{\"u}rnkranz, J.
	\newblock A survey of preference-based reinforcement learning methods.
	\newblock \emph{The Journal of Machine Learning Research}, 18\penalty0
	(1):\penalty0 4945--4990, 2017.
	
\end{thebibliography}
\end{document}